\pdfoutput=1
\documentclass[11pt]{article}

\usepackage[final]{acl}

\usepackage{times}
\usepackage{latexsym}
\usepackage[T1]{fontenc}
\usepackage[utf8]{inputenc}
\usepackage{microtype}
\usepackage{inconsolata}
\usepackage{graphicx}
\usepackage[most]{tcolorbox}
\usepackage{color, colortbl}
\definecolor{Gray}{gray}{0.9}
\usepackage{tabularx}
\usepackage{arydshln}
\usepackage{multirow}
\usepackage{enumitem}
\usepackage[normalem]{ulem}
\usepackage{graphicx}
\usepackage{subcaption}
\usepackage{hyperref}
\usepackage{url}
\usepackage{cleveref}
\usepackage{booktabs}
\usepackage{xspace}
\usepackage{amssymb}
\usepackage{footnotehyper}
\usepackage{algorithm}
\usepackage{algpseudocode}
\usepackage{array}
\usepackage{xcolor}
\usepackage{colortbl}

\newcommand{\Dname}{\textsc{SciDQA}\xspace}

\newcommand{\closedbook}{\texttt{closed-book}\xspace}
\newcommand{\titleabs}{\texttt{title-abs}\xspace}
\newcommand{\fulltext}{\texttt{full-text}\xspace}

\title{\Dname: A Deep Reading Comprehension Dataset over Scientific Papers}

\author{Shruti Singh$^1$ \quad Nandan Sarkar$^2$ \quad Arman Cohan$^{2,3}$ \\
$^1$IIT Gandhinagar \quad $^2$Yale University \quad $^3$Allen Institute for AI
  }

\begin{document}
\maketitle
\begin{abstract}
Scientific literature is typically dense, requiring significant background knowledge and deep comprehension for effective engagement. 
We introduce \Dname, a new dataset for reading comprehension that challenges LLMs for a deep understanding of scientific articles, consisting of 2,937 QA pairs. Unlike other scientific QA datasets, \Dname sources questions from peer reviews by domain experts and answers by paper authors, ensuring a thorough examination of the literature. We enhance the dataset's quality through a process that carefully filters out lower quality questions, decontextualizes the content, tracks the source document across different versions, and incorporates a bibliography for multi-document question-answering.
Questions in \Dname necessitate reasoning across figures, tables, equations, appendices, and supplementary materials, and require multi-document reasoning.
We evaluate several open-source and proprietary LLMs across various configurations to explore their capabilities in generating relevant and factual responses. Our comprehensive evaluation, based on metrics for surface-level similarity and LLM judgements, 
highlights notable performance discrepancies. \Dname represents a rigorously curated, naturally derived scientific QA dataset, designed to facilitate research on complex scientific text understanding.
\end{abstract}

\section{Introduction}
Question-answering (QA) datasets are valuable for evaluating the reading comprehension, reasoning, and document understanding capabilities of language models~\citep{dua-etal-2019-drop,dasigi2021dataset,rogers2023qa}.
The scientific QA task involves reading a research paper and answering questions, drawing on the paper content and some background knowledge. This task mirrors how humans engage with academic literature \cite{Lo2023TheSR,Palani2023RelatedlySL}.

Scientific literature is inherently dense and typically requires a deep understanding and significant background knowledge to fully comprehend and engage with. To address this challenge, the NLP community has developed various datasets for question-answering (QA) from research papers to aid in development and evaluation of AI systems for comprehending the research papers. Methods range from manual question generation by domain experts~\cite{moller-etal-2020-covid,dasigi2021dataset,lee2023qasa} to automated extraction of questions using machine learning from selected texts~\citep{saikh2022scienceqa,saikh-etal-2020-scholarlyread,pappas-etal-2020-biomrc,jin-etal-2019-pubmedqa,pappas-etal-2018-bioread}. However, many of these datasets focus on surface-level information and are often limited to questions that are written from titles and abstracts, which restricts the complexity and deeper engagement with the full papers.

\begin{figure}[t]
    \centering
    \includegraphics[width=0.98\linewidth]{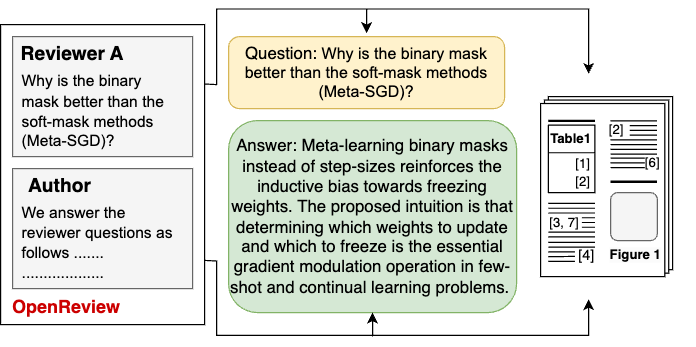}
    \caption{An instance in the SciDQA dataset. The question and answer corresponding to the paper are extracted from the reviewer-author discussion on OpenReview.}
    \label{fig:sampleQA}
\end{figure}

\begin{figure}[t]
    \centering
    \includegraphics[width=0.98\linewidth]{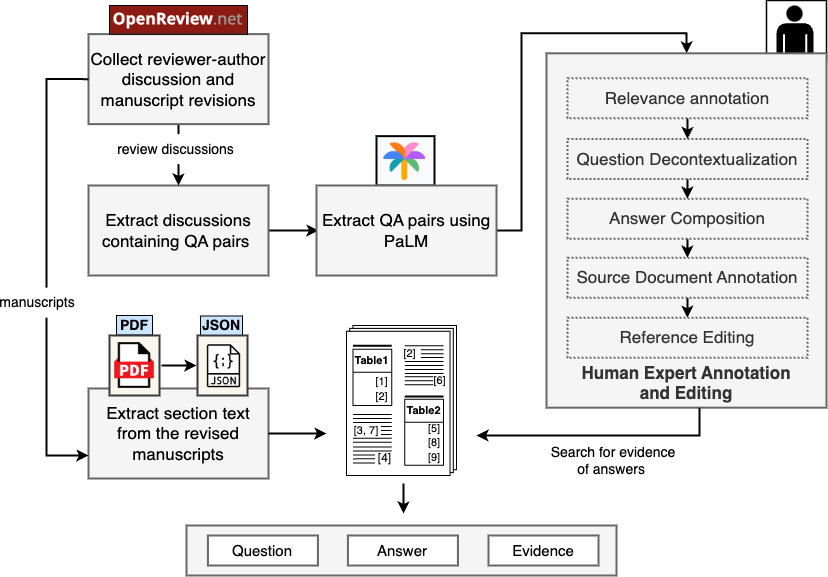}
    \caption{Dataset curation pipeline for \Dname. LLM-based QA extraction from peer reviews is followed by a comprehensive human expert annotation and editing. As discussed, we only include evidence for a subset of the dataset due to high annotation cost.}
    \label{fig:data_curation_pipeline}
\end{figure}

We introduce \Dname, a novel deep reading comprehension dataset for scientific papers.
It is specifically tailored to the scientific articles in the machine learning (ML) domain and sourced from peer reviews on the OpenReview platform~\citep{openreview}.
Peer reviews frequently include questions or comments from reviewers who seek information or clarification on aspects they are confused about or do not fully understand. Answering many of such questions necessitate a deep and comprehensive understanding of the research and the background, such as a critical view of the approach and results, implications of the findings, and comparisons with previous works. Moreover, peer reviews are acompanied by responses from authors, who have carefully tried to address and clarify the reviewers questions.   
As both authors and reviewers are domain experts, responding to these inquiries necessitates a deep understanding of the paper and its broader research field. 
Consequently, we believe these questions are an excellent source for probing deep comprehension of research papers, contrasting with prior work that often targets shallow information-extraction or surface-level facts.
However, not all such questions expressed in a review are useful. In addition they also need rewriting to stand alone as clear, self-contained queries suitable for a reading comprehension dataset. To ensure the quality and relevance of our dataset, we implement a human annotation process by domain experts, highlighted in~\Cref{fig:data_curation_pipeline}.

Our dataset features long-form questions and answer pairs, as shown in~\Cref{tab:data_stats}.
It is diverse and also some questions require comprehension of figures, tables, equations, and references in addition to the paper text. Approximately 11\% of the questions necessitate reasoning over at least one explicitly mentioned reference paper in addition to the candidate paper. 
We evaluate several open-source and proprietary LLMs under various configurations (including closed-book, retrieval-based setup and long-context reasoning) to benchmark their capabitlies on this task. 
Our findings suggest that our dataset presents a significant challenge, as several LLMs struggle to generate accurate factual answers across a variety of experimental setups. Our dataset, code, and model outputs to reproduce our results are available on the github repository.~\footnote{\href{https://github.com/yale-nlp/SciDQA}{https://github.com/yale-nlp/SciDQA}}

\newcolumntype{B}{>{\centering\arraybackslash} p{0.1\linewidth}}
\begin{table*}[!htbp]
\begin{center}
\small{
\begin{tabular}{@{}p{0.15\linewidth}lrlBBBB@{}}
\toprule
Dataset       & Curation & Size & Source & Question length & Answer length & Multiple Docs & \% Short Answers \\
\midrule
QASA~\citeyearpar{lee2023qasa}      &  Manual &  1,554  & Full-Text   &  15.86  &  44.95  & $\times$ & 1.61\% \\
QASPER~\citeyearpar{dasigi2021dataset}     & Manual &  5,089  & Title/Abstract   &  9.33   &  18.19  & $\times$ & 39.94\% \\
Covid-QA~\citeyearpar{moller-etal-2020-covid}   & Manual &  2,019  & Full-Text   &  10.61       & 15.79  & $\times$    & 32.64\%  \\
ScholarlyRead$^\dag$ ~\citeyearpar{saikh-etal-2020-scholarlyread} & Synthetic &  10,000  & Abstract   &    NA     &   NA   & $\times$  & NA  \\
BioRead \citeyearpar{pappas-etal-2018-bioread}     &  Synthetic & 16.4M  &  Full-Text     &    42.90   &    1.92   & $\times$  & 98.70\% \\
BioMRC~\citeyearpar{pappas-etal-2020-biomrc}    & Synthetic  &  700,000  & Title/Abstract   &  16.01       &   1.73    & $\times$ & 99.38\% \\
PubMedQA \citeyearpar{jin-etal-2019-pubmedqa} & & & & & & \\
\hspace{0.3cm} Annotated  &  Manual & 1,000   &  Title/Abstract  &   14.42     &  43.23     & $\times$  & 0\%\textsuperscript{*} \\
\hspace{0.3cm} Unlabeled  & Synthetic  & 61,249   & Title/Abstract   & 14.98      &  45.88   & $\times$ & 0\%\textsuperscript{*} \\
\hspace{0.3cm} Artificial & Synthetic & 211,269  & Title/Abstract   &   16.35  &  40.97  & $\times$ & 0\%\textsuperscript{*} \\ \hdashline
\Dname (Ours)      & Hybrid & 2,937   & Full-Text &  23.92  &  104.67 & \checkmark & 1.74\% \\
\bottomrule
\end{tabular}
}
\end{center}
\caption{
Comparison of the related datasets.
$^\dag$ScholarlyRead dataset is unavailable publicly, hence we skip its statistics.
\textsuperscript{*}PubMedQA features two types of answers: a long answer, which is the last sentence of the abstract, and a short answer, which is yes/no. Here, we report statistics of long answers as all short answers are less than 5 words.}
\label{tab:data_stats}
\end{table*}

\section{Building the \Dname~Dataset}
We present the pipeline for the collection of the  \Dname~dataset and the preprocessing, manual filtering and rewriting steps involved. A schematic denoting the pipeline is presented in~\Cref{fig:data_curation_pipeline}. We present the various stages of data curation next.

\subsection{Curation from OpenReview}
We selected top-tier ML and DL venues, designated as A* rankings by ICORE Portal \citep{icoreranking}, with publicly accessible reviewer-author discussions on OpenReview (Appendix \ref{app:preproc}). We curate 11400 papers from ICLR (2018-2022) and NeurIPS (2021-2022), with a major focus on including newer papers to decrease the risk of contamination with LLM pretraining datasets. 

\subsection{Processing the reviews}

\paragraph{PDF to Text Conversion}
OpenReview portal hosts multiple submitted PDF versions of a submitted manuscript which are curated. Nougat~\citep{blecher2023nougat}, a visual transformer model designed for scientific OCR tasks (details in Appendix~\ref{app:preproc}), is used for PDF to text conversion.

\paragraph{Regex Filtering}
OpenReview has nested discussions, i.e. authors and reviewers reply to messages, creating a time-stamp chain of discussion. We extract 18,658 reviewer-author discussions for 11,400 papers that contain questions and answers, by regex pattern matching (details in Appendix~\ref{app:preproc}). 

\paragraph{LLM-based QA Extraction}
Next, we extract explicit questions that reviewers asked the authors from the reviews. For QA extraction, we utilized the PaLM API \citep{palmtbapi} to extract specific question-answer pairs within the reviewer-author discussions.\footnote{We chose to use PaLM because it consistently delivered high-quality extractions and offered an available API, capable of handling up to 60 requests per minute.} Initial attempts to extract questions and answers using non-LLM methods faced challenges, as authors and reviewers employ various patterns for posing questions and answers, making it difficult to comprehensively cover all instances. Through this approach, we extracted 26,085 question-answer pairs. Details of the prompts are in the Appendix~\ref{app:preproc}~\Cref{fig:PALMExtractionPrompt}.

\subsection{Human Expert Annotation and Editing}
In initial investigations, we found that many of the extracted questions are \textit{not} useful and they would need additional revisions to be appropriate for a QA dataset.
Therefore, to ensure the quality of the QA pairs in the \Dname~dataset, we employed an extensive manual annotation process by domain experts.\footnote{Students with extensive experience in NLP and ML.} This included determining and keeping only the most relevant questions, rewriting both questions and answers, and editing references in the QA pairs. We briefly discuss annotation and editing stages.
\paragraph{Relevance Annotation}

This task selects information-seeking questions, whose answers are identifiable within the research paper text, from a set of synthetically generated QA pairs. Questions referencing figures, tables, equations, specific sections, or lines, and inquiries requiring data from multiple papers were categorized as relevant. Conversely, questions asking for edits, summaries, or subjective judgments about the paper's quality, or those based on the authors' personal experiences, were classified as irrelevant. To expedite the annotation process, we introduced an ‘ambiguous’ category for cases where the relevance of a question-answer pair was challenging to ascertain. Questions necessitating experimental validation for answers, and where it remained unclear whether the authors had conducted such experiments based on reviewer suggestion during reviewer-author discussion, were classified as ambiguous. We present a few samples for each category in~\Cref{tab:cat_instances} in Appendix~\ref{app:preproc}. 

Two annotators, also the authors of this paper, annotated the dataset, starting with a common subset of 200 instances and achieving an 85\% agreement rate. The disagreements were discussed and resolved, and the rest of the questions were annotated by a single annotator.
In total, the annotators reviewed 7,000 instances, identifying 2,937 QA pairs as relevant, equivalent to a relevancy rate of approximately 41\%. 
Additional details about the annotations are in Appendix~\ref{apx:annotation}.

\paragraph{Decontextualizing Questions and Answers}
Originally, questions were directed towards the authors of the paper and authors provided answers from their perspective. We rewrote these QA pairs in the third-person point of view to make them universally applicable and to avoid biasing language models to generate answers in the first person when trained on \Dname. This is also necessary for the models to understand that the question does not ask for their personal opinion, but is a factual question seeking information about the author's reasoning in the paper. We also add contextual information to the questions where the question is incomplete or incomprehensible without contextual information present in the review text. We present an example in~\Cref{fig:editing_qa} showcasing scenarios where decontextualization and editing the narrative is necessary to comprehend the question. The perplexity of questions before and after rewriting, when evaluated with the GPT-2 model, exhibits a difference of 16.3 points, suggesting that decontextualization contributes to an enhancement in dataset quality.
\paragraph{Annotating the Source Document}
Certain conferences like NeurIPS and ICLR allow authors to submit revised manuscripts during the author-reviewer discussion period. For simplicity, we focus only on the initial submitted copy and the final camera-ready manuscript. For rejected papers, the last submitted manuscript is considered the final version, which may sometimes be identical to the initial submission. Establishing the source document between the initial and final manuscripts presents challenges, as author-reviewer discussions often result in added details like tables, figures, and text, making the camera-ready version a suitable source document. However, reviewers' questions may prompt authors to rewrite paper text to explicitly mention the answer, simplifying the dataset if the final version is used.  We depict two such scenarios in~\Cref{fig:initial_vs_camready}. To manage these variations, each question-answer pair is annotated with the version of the document used as the source, typically the initial or final version. If author responses indicate additions in a revision, the final version is marked as the source document. If no specific information is given, the initial version is defaulted as the source. This approach addresses potential ambiguities arising from updates in table, figure, and section numbers in the revised final manuscript.

\paragraph{Reference Editing}
Finally, to prevent language models from taking shortcuts by extracting answers based on reference text markers within the papers, we edited the references in the QA pairs, as shown in~\Cref{fig:ref_editing}. This process involved replacing specific reference markers with placeholders and providing a list of necessary references at the end of the question and the answer. 

\section{Dataset Details and Analysis} 
The \Dname~dataset comprises 2,937 question-answer pairs. We present the statistics of \Dname~in comparison to other related existing QA datasets in~\Cref{tab:data_stats}. 
Next, we discuss the diversity of answer sources, and fuzzy searching for answers, and the statistics of changes in initial and revised manuscripts.

\paragraph{Diversity of Answer Sources}
Our dataset features questions necessitating reasoning across multiple modalities beyond mere text, including figures, tables, equations, and both appendix and supplementary materials.\footnote{For experiments, we use table and figure captions and do not use multi-modal models for direct processing of figures. We'll leave that as a future direction.} This design ensures that comprehensive reasoning over the full-text of the paper is essential for answering the questions accurately. The statistics are presented in~\Cref{tab:info_source}.
\begin{table}[]
\begin{center}
\small{
\begin{tabular}{p{0.6\linewidth} p{0.3\linewidth}}
\toprule
\rowcolor{Gray}
Information Source         & \% in Dataset \\
\midrule
Tables                     & 14.03\%               \\
Multiple documents         & 10.9\%                \\
Appendix and Supplementary & 10.01\%               \\
Equations and Symbols      & 10.32\%               \\
Figures                    & 6.98\% \\
\bottomrule
\end{tabular}
}
\end{center}               
\caption{Distribution of various modalities (text, figures, tables, equations, appendix, and supplementary) which are required to answer the questions in the dataset.}
\label{tab:info_source}
\end{table}

\paragraph{Fuzzy Search for Answers}
\label{ssec:ans_evidence}
We search for answers in the research paper texts and find sections with at least 80\% unigram overlap between answers and paragraphs. Such a high degree of overlap suggests that the text from the research papers is directly utilized as answers to questions, simplifying the question-answering process to the identification of pertinent paragraphs. This implies a reduced necessity for reasoning or inferential thinking compared to scenarios where answers must be derived from an analysis of the text. Our findings reveal that only 25\% of the answers in our dataset can be identified with an overlap exceeding 80\%. By contrast, the QASA dataset \cite{lee2023qasa}, features 52\% of answers that demonstrate more than 80\% unigram overlap with the paper text, indicating a higher reliance on direct text retrieval for answering questions. 

\paragraph{Edits in Initial and Revised Manuscripts}
\label{ssec:manuscript_version}
We conducted an analysis of differences between PDF versions
for each QA pair.\footnote{This is because authors often update their manuscripts in response to comments and questions by reviewers.} 
Our dataset of 576 unique papers shows that 66.3\% vary in figure mentions, and 54.9\% vary in table counts between initial and final manuscripts, highlighting the need to maintain separate versions.

\section{Experimental Setup}
We design four task configurations to evaluate the capabilities of LLMs in answering the questions in~\Dname. We use two separate setups, closed-book \cite{roberts-etal-2020-much}, and open-book. We experiment with a wide-range of open-source LLMs (Falcon~\citep{falconmodel}, Galactica~\citep{taylor2022galactica}, Gemma~\citep{team2024gemma}, Llama 2~\citep{llama2model}, Llama 3.1~\citep{llama31model} Mistral~\citep{jiang2023mistral}, Phi-2~\citep{2023phi2}, Qwen v2.5~\citep{qwen2.5}, Vicuna~\citep{vicuna2023}, and Zephyr~\citep{zephyrmodel}) and two frontier closed models Gemini Pro~\citep{team2023gemini} and GPT-4~\citep{achiam2023gpt} models. For open-source models, we experiment with various model sizes from $\sim$2B to $\sim$70B parameters.

\paragraph{Priming with the Question Only (\closedbook)}
Can LLMs answer the questions in a closed-book setting \cite{roberts-etal-2020-much} when primed only with the question text and without explicitly providing the paper? LLMs have the ability to retain knowledge and in this sense, it's conceivable that LLMs might be able to generate answers directly based solely on the question text, without any context from the associated research papers.\footnote{Comphrehensive evaluation of this setting is challenging, as it is difficult to disentangle potential effect of contamination, from knowledge retrained by LLMs, especially in models where source of training data isn't disclosed. While we source our questions from peer reviews, our questions and answers are significantly revised and re-written, so exact-match contamination is less likely.} Further, LLMs might already have internalized the knowledge related to papers to be able to answer some specific questions without explicitly providing the context. To investigate this possibility, in the \closedbook configuration, LLMs are presented with only the questions and instructions, without any information about the relevant paper.

\paragraph{Priming with Question, and Paper's Title and Abstract (\titleabs)}
In this setting, we provide the LLM with the question text, along with the Title/Abstract of the paper. This mimicks a ``partially'' \closedbook setting. The objective is to ascertain whether the inclusion of limited additional information, such as the paper's Title and Abstract, enhances the LLM's ability to accurately retrieve and recall the knowledge to correctly answer the question. Unlike the fully \closedbook setting, it is not entirely infeasible to answer some questions with the information provided in the abstract. However, given that our dataset comprises questions that require complex reasoning, the answers to the majority of questions will not be found in the abstract alone. 

\paragraph{Retrieval-Augmented Generation with LLMs (RAG)}
We follow a retrieval-augmented generation setup for this configuration. Research paper texts exceed the typical model context length with exception of few long-context models (which we will discuss in the next experimental setup).
To accomodate processing such documents we employ a RAG setup, where we first divide the document into smaller and slightly overlapping chunks, retrieve the most relevant chunks to the question using a BM25 ranker,\footnote{More advanced retrieval settings using dense retrievers or rerankers can be also employed to improve the performance of models in this setting. Our goal is mainly to provide a baseline setup for each of the experimental settings.} and subsequently input the top ranked chunks to the LLM, tasked with generating the response.
The operational flow of this pipeline is depicted in Appendix~\Cref{fig:rag} and the chunking algorithm is presented in Appendix~\Cref{alg:chunking_rag}.

\paragraph{Comprehending the Full-text using LLMs (\fulltext)}
In this experimental setup, LLMs are provided with the full-text of scientific papers and are tasked with answering a specific question. The length of scientific texts could exceed the context length limit of many LLMs. In such cases, we divide the full-text into segments. Each segment, along with the question and instructions, is then presented to an LLM (referred to as base-LLM), which generates answers for each segment.

This setup produces multiple answer candidates for a single question, contingent on the number of passes required to present all chunks to the LLM. To distill these into a singular, optimal response, we introduce an answer selection phase. During this phase, the Llama 3.1 70B model is prompted with the question and all answers generated by the base-LLM, with instructions to identify the most comprehensive response from the provided options. Details of this prompt are included in the Appendix~\ref{app:exp_details} in~\Cref{fig:answer_selection_prompt}. We only segment paper's full-text when it exceeds the model's context length (pipeline presented in Appendix~\Cref{fig:ft}). 

For models with context length limit greater than the full-text (Gemini, GPT-4o, and GPT-4o-mini), the base-LLM directly generates the answer from the full text, and the answer-selection phase is not required. For Qwen v2.5 (1.5B and 7B) and Llama 3.1 (8B and 70B) models, the context length is 128k, however the prompt with the entire paper text does not fit into the cache, so we chunk the text and generate multiple answer candidates similar to other LLMs, however, the answer selection phase uses the same base-LLM (Qwen and Llama versions respectively) instead of Llama 3.1 70B.

\subsection{Evaluation}
\paragraph{Surface-level Metrics:} We first use surface-level metrics for evaluating the LLM generated answers, which compare the similarity of the generated long-form answer with the gold standard through textual overlaps. These include ROUGE score~\citep{lin-2004-rouge} (we compute ROUGE-1 (R-1), ROUGE-2 (R-2), and ROUGE-L (R-L); and report the average as $\texttt{R}_\mu$), BLEURT-20~\citep{pu-etal-2021-learning} (abbreviated as \texttt{BL}), and BERTScore~\citep{bert-score} (BERTScore F1 score as \texttt{BS}). 

\begin{table}[!t]
\begin{center}
\renewcommand{\arraystretch}{1.1}
\small{
\begin{tabular}{@{}lcccc@{}}
\toprule
\rowcolor{Gray}
\multirow{1}{*}{\textbf{Model}} & \textbf{CB} & \textbf{T/Abs} & \textbf{RAG} & \textbf{FT} \\ 
\midrule
\multicolumn{5}{c}{2-3 B} \\ 
\midrule
Gemma IT \citeyearpar{team2024gemma} & 40.75 & 31.50 & 39.47 & 30.33 \\
Phi2 \citeyearpar{2023phi2} & 45.24 & 43.16 & 42.20 & 40.95 \\
Qwen 2.5 IT \citeyearpar{qwen2.5} & 34.86 & 37.81 & 33.01 & 35.70 \\
\midrule
\multicolumn{5}{c}{6-7 B} \\ 
\midrule
Falcon IT \citeyearpar{falconmodel} & 28.49 & 19.70 & \cellcolor[HTML]{E8F0F8}44.28 & \cellcolor[HTML]{E8F0F8}42.25 \\
Galactica \citeyearpar{taylor2022galactica} & 14.49 & 41.76 & 34.27 & 43.07 \\
Llama 2 Chat \citeyearpar{llama2model} & 26.09 & 36.20 & \cellcolor[HTML]{E8F0F8}46.95 & \cellcolor[HTML]{E8F0F8}45.99 \\
Llama 3.1 IT \citeyearpar{llama31model} & 20.46 & 42.56 & 38.59 & \cellcolor[HTML]{E8F0F8}45.73 \\
Longchat 32k \citeyearpar{longchat2023} & 25.77 & 22.59 & \cellcolor[HTML]{E8F0F8}44.98 & \cellcolor[HTML]{E8F0F8}40.58 \\
Mistral IT \citeyearpar{jiang2023mistral} & 29.71 & 47.64 & 47.67 & 42.29 \\
Qwen 2.5 IT \citeyearpar{qwen2.5} & 44.72 & 47.10 & 45.13 & 41.41 \\
Vicuna \citeyearpar{vicuna2023} & 21.32 & 18.22 & \cellcolor[HTML]{E8F0F8}42.01 & \cellcolor[HTML]{E8F0F8}46.46 \\
Zephyr $\beta$ \citeyearpar{zephyrmodel} & 29.20 & 41.66 & \cellcolor[HTML]{E8F0F8}\textbf{48.74} & 42.13 \\
\midrule
\multicolumn{5}{c}{13 B} \\ 
\midrule
Llama 2 Chat \citeyearpar{llama2model} & 28.06 & 37.35 & \cellcolor[HTML]{E8F0F8}47.53 & \cellcolor[HTML]{E8F0F8}45.88 \\
Vicuna \citeyearpar{vicuna2023} & 27.69 & 30.11 & \cellcolor[HTML]{E8F0F8}45.41 & \cellcolor[HTML]{E8F0F8}46.77 \\
\midrule
\multicolumn{5}{c}{70 B} \\ 
\midrule
Llama 2 Chat \citeyearpar{llama2model} & 43.33 & 39.69 & 40.71 & 30.14 \\
Llama 3.1 IT \citeyearpar{llama31model} & 46.20 & 48.46 & 47.60 & 47.78 \\
\midrule
\multicolumn{5}{c}{Proprietary LLMs} \\ 
\midrule
Gemini Pro \citeyearpar{team2023gemini} & 28.31 & 32.01 & \cellcolor[HTML]{E8F0F8}38.03 & \cellcolor[HTML]{E8F0F8}37.59 \\
GPT-4o & \textbf{48.48} & \textbf{50.61} & 46.63 & \cellcolor[HTML]{E8F0F8}\textbf{54.03} \\
GPT-4o-mini & 47.50 & \textbf{50.32} & \textbf{48.90} & \cellcolor[HTML]{E8F0F8}\textbf{54.02} \\
GPT-4o \citeyearpar{achiam2023gpt} & - & - & - & 49.3 \\
\bottomrule
\end{tabular}
}
\end{center}
\caption{Average scores for all configurations. CB refers to \closedbook and T/Abs referes to \titleabs. Cells in blue indicate RAG or \fulltext (FT) settings where performance improves over both \closedbook settings by atleast two points. For seven models, both RAG and \fulltext lead to better scores, while for four models (including GPT-4o and GPT-4o-mini) only one of the RAG/\fulltext settings performs significantly better than Closed settings.}
\label{tab:avg_score_comparison}
\end{table}

\paragraph{LLM Judge:} In addition to traditional surface-level metrics, we also use LLM-as-a-judge to evaluate the quality of the generated text, given potential unreliability issues of surface-level metrics \cite{liu-etal-2023-revisiting}. In particular, we employ Llama 3.1 70B~\citep{llama31model}, GPT-4o, and GPT-4o-mini\footnote{Llama 3.1 allows full reproducible results, and is a highly capable open-source model in evaluating instruction-following \cite{liu2024reife}, GPT-4o is a frontier LLM at time of writing, and GPT-4o-mini balances the cost and quality of evaluation.} to evaluate LLM-generated answers on four aspects, namely relevance, accuracy, completeness, and conciseness on a scale of 1-10.
The LLM is also asked to report the overall quality scores by averaging the individual scores for each aspect. All LLM judge models (Llama 3.1 70B, GPT-4o, and GPT-4o-mini) are prompted to generate the explanations and the scores for each aspect, and subsequently, Llama 3.1 8B model is used to extract the overall quality score from the generated explanations (see appendix \ref{appx:llm-judge-prompts} for the exact prompts used). The average scores, normalized to range 1-100, from Llama 3.1 70B, GPT-4o, and GPT-4o-mini judges are represented with \texttt{L70}, \texttt{4o}, and \texttt{4oM} respectively. The average of LLM-as-a-judge scores is presented as \texttt{ALS} in the tables. An average score of all traditional and LLM judge scores is presented in column \texttt{Avg}.

\section{Results and Discussion}

\newcolumntype{a}{>{\columncolor{blue!7}}c}
\newcolumntype{b}{>{\columncolor{green!10}}c}

\begin{table*}[!t]
\renewcommand{\arraystretch}{1.1}
\begin{center}
\addtolength{\tabcolsep}{-0.3em}
\small{
\begin{tabular}{@{}ll:ccccccaa:ccccccaa@{}}
\toprule
\rowcolor{Gray}
& & \multicolumn{8}{c}{\textbf{\closedbook (Prompt \textless{}INS, Q\textgreater{})}} & \multicolumn{8}{c}{\textbf{\titleabs (Prompt \textless{}INS, Q, TABS\textgreater{})}} \\ 
\rowcolor{Gray}
\textbf{Size} & \textbf{Model} & $\textbf{\texttt{R}}_{\mu}$ & \textbf{\texttt{BL}} & \textbf{\texttt{BS}} & \textbf{\texttt{L70}} & \textbf{\texttt{4o}} & \textbf{\texttt{4oM}} & \textbf{\texttt{ALS}} & \textbf{\texttt{Avg}} & $\textbf{\texttt{R}}_{\mu}$ & \textbf{\texttt{BL}} & \textbf{\texttt{BS}} & \textbf{\texttt{L70}} & \textbf{\texttt{4o}} & \textbf{\texttt{4oM}} & \textbf{\texttt{ALS}} & \textbf{\texttt{Avg}} \\ 
\cmidrule(r){1-2} \cmidrule(lr){3-10} \cmidrule(l){11-18}
\multirow{3}{*}{2-3 B} & Gemma IT \citeyearpar{team2024gemma} & 12.3 & 41.1 & 50.0 & 46.9 & 43.5 & 50.6 & 47.0 & 40.8 & 15.7 & 29.5 & 47.2 & 33.6 & 30.8 & 32.3 & 32.2 & 31.5 \\
& Phi-2 \citeyearpar{2023phi2} & 16.4 & 40.8 & 54.1 & 53.4 & 49.3 & 57.4 & 53.4 & 45.2 & 12.3 & 42.8 & 50.6 & 49.1 & 48.8 & 55.3 & 51.1 & 43.2 \\
& Qwen 2.5 IT \citeyearpar{qwen2.5} & 5.9 & 53.1 & 42.5 & 31.7 & 32.3 & 43.7 & 35.9 & 34.9 & 6.3 & 54.1 & 42.4 & 37.8 & 37.5 & 48.8 & 41.4 & 37.8 \\
\cmidrule(r){1-2} \cmidrule(lr){3-10} \cmidrule(l){11-18}
\multirow{7}{*}{6-7 B} & Falcon IT \citeyearpar{falconmodel} & 14.7 & 0.4 & 0.5 & 53.2 & 46.4 & 55.7 & 51.8 & 28.5 & 4.9 & 9.5 & 33.9 & 30.8 & 18.4 & 20.7 & 23.3 & 19.7 \\
& Galactica \citeyearpar{taylor2022galactica} & 4.1 & 0.5 & 0.4 & 23.2 & 26.6 & 32.2 & 27.3 & 14.5 & 20.1 & 40.3 & 49.9 & 45.5 & 44.5 & 50.3 & 46.8 & 41.8 \\
& Llama 2 Chat \citeyearpar{llama2model} & 7.5 & 0.5 & 0.5 & 46.8 & 46.2 & 55.1 & 49.4 & 26.1 & 10.6 & 37.1 & 48.9 & 40.2 & 37.5 & 42.9 & 40.2 & 36.2 \\
& Llama 3.1 IT \citeyearpar{llama31model} & 1.7 & 11.0 & 32.8 & 33.4 & 21.1 & 22.8 & 25.8 & 20.5 & 6.1 & 46.4 & 45.8 & 48.3 & 51.7 & 57.1 & 52.4 & 42.6 \\
& Longchat 32k \citeyearpar{longchat2023} & 9.4 & 0.4 & 0.5 & 51.4 & 41.8 & 51.1 & 48.1 & 25.8 & 1.3 & 9.3 & 34.1 & 59.4 & 12.7 & 18.8 & 30.3 & 22.6 \\
& Mistral IT \citeyearpar{jiang2023mistral} & 13.7 & 0.4 & 0.5 & 55.4 & 50.2 & 58.0 & 54.5 & 29.7 & 16.3 & 40.3 & 53.5 & 59.3 & 56.0 & 60.4 & 58.6 & 47.6 \\
& Qwen 2.5 IT \citeyearpar{qwen2.5} & 6.5 & 46.2 & 46.3 & 54.2 & 54.1 & 61.0 & 56.4 & 44.7 & 7.5 & 45.1 & 47.4 & 60.7 & 59.0 & 62.9 & 60.9 & 47.1 \\
& Vicuna \citeyearpar{vicuna2023} & 7.0 & 0.3 & 0.4 & 53.1 & 29.9 & 37.2 & 40.1 & 21.3 & 2.8 & 5.9 & 31.3 & 29.5 & 23.4 & 16.5 & 23.1 & 18.2 \\
& Zephyr $\beta$ \citeyearpar{zephyrmodel} & 9.5 & 0.4 & 0.5 & 54.6 & 51.0 & 59.1 & 54.9 & 29.2 & 13.0 & 38.4 & 51.2 & 49.2 & 46.5 & 51.7 & 49.1 & 41.7 \\
\cmidrule(r){1-2} \cmidrule(lr){3-10} \cmidrule(l){11-18}
\multirow{2}{*}{13 B} & Llama 2 Chat \citeyearpar{llama2model} & 7.7 & 0.5 & 0.5 & 51.5 & 50.1 & 58.1 & 53.2 & 28.1 & 10.7 & 39.1 & 49.6 & 41.3 & 39.8 & 43.7 & 41.6 & 37.4 \\
& Vicuna \citeyearpar{vicuna2023} & 9.3 & 0.4 & 0.5 & 54.2 & 47.5 & 54.2 & 52.0 & 27.7 & 7.9 & 22.5 & 41.1 & 40.7 & 32.6 & 35.8 & 36.4 & 30.1 \\
\cmidrule(r){1-2} \cmidrule(lr){3-10} \cmidrule(l){11-18}
\multirow{2}{*}{70 B} & Llama 2 Chat \citeyearpar{llama2model} & 8.2 & 44.7 & 49.0 & 50.2 & 49.8 & 58.0 & 52.7 & 43.3 & 6.2 & 49.5 & 43.7 & 40.2 & 46.7 & 51.8 & 46.2 & 39.7 \\
& Llama 3.1 IT \citeyearpar{llama31model} & 10.5 & 44.2 & 50.0 & 57.0 & 55.4 & 60.1 & 57.5 & 46.2 & 12.7 & 42.7 & 52.1 & 61.3 & 60.0 & 62.0 & 61.1 & 48.5 \\
\cmidrule(r){1-2} \cmidrule(lr){3-10} \cmidrule(l){11-18}
\multirow{3}{*}{UNK} & Gemini Pro \citeyearpar{team2023gemini} & 5.3 & 21.1 & 39.1 & 38.9 & 31.7 & 33.8 & 34.8 & 28.3 & 12.0 & 24.6 & 44.2 & 40.0 & 36.6 & 34.7 & 37.1 & 32.0 \\
& GPT-4o & 9.0 & 42.6 & 49.1 & 64.3 & 60.8 & 65.1 & \textbf{63.4} & \textbf{48.5} & 11.7 & 42.3 & 51.5 & 66.7 & 64.9 & 66.6 & 66.1 & \textbf{50.6} \\
& GPT-4o-mini-2 & 7.7 & 42.9 & 48.5 & 62.4 & 59.3 & 64.3 & 62.0 & 47.5 & 9.1 & 43.2 & 50.1 & 67.1 & 64.9 & 67.5 & \textbf{66.5} & 50.3 \\ \bottomrule
\end{tabular}
}
\end{center}
\caption{\closedbook evaluates LLMs in a closed book setting without the paper context. \titleabs configuration evaluates if additional context (title and abstract) helps in answering the questions. The metrics are $\texttt{R}_\mu$ (average of rouge-1, rouge-2, and rouge-l), BLEURT-20 (\texttt{BL}), BERTScore F1 (\texttt{BS}), and LLM judges Llama 3.1 70B (\texttt{L70}), GPT-4o (\texttt{4o}), and GPT-4o-mini-2 (\texttt{4oM}). \texttt{ALS} is the average over LLM judges, and \texttt{Avg} is the average over all metrics.}
\label{tab:odqa_results}
\end{table*}

In this section we dicsuss the main results. We provide the main results as averages of all metrics for all the four experimental settings in Table \ref{tab:avg_score_comparison}. Then we present the detailed results of each setting in Tables \ref{tab:odqa_results} and \ref{tab:ragcft_results}.

\paragraph{Comparison among \closedbook, \titleabs, RAG, and \fulltext settings:}
We compare the average scores of all metrics for the models in four settings in~\Cref{tab:avg_score_comparison}, out of which two settings (\closedbook and \titleabs) are open-domain question answering settings, and the other two settings provide the LLMs with paper context either in relevant chunk format (RAG) or full-text of the paper (\fulltext). Seven models (Falcon Instruct 7B, Llama 2 Chat 7B and 13B, Longchat 7B, Vicuna 7B and 13B, and Gemini) perform better when provided with paper context (both RAG and \fulltext setup) over the other two open-domain question answering settings (Priming with Questions (\closedbook) and Priming with Question and Title/Abstract (\titleabs)). Other four models (Llama 3.1 8B, Zephyr 7B, GPT-4o, and GPT-4o-mini) perform better only in one of the RAG/\fulltext settings in comparison to open-domain question answering settings. The rest seven models show a degradation or similar scores when provided with paper context, likely indicating that either there could be contamination affecting the results or the models are able to generate shallow answers without reasoning about the question.

We compute the maximum improvement in scores when provided with paper context (\fulltext setting), by computing the difference of best scores in \closedbook and RAG/\fulltext settings. Vicuna 7B model shows the highest improvement in average scores (25 points) from \closedbook to \fulltext setting, indicating it is able to effectively use the papers full-text to reason about the question.  

\begin{table*}[!t]
\renewcommand{\arraystretch}{1.1}
\begin{center}
\addtolength{\tabcolsep}{-0.3em}
\small{
\begin{tabular}{@{}ll:ccccccaa:ccccccaa@{}}
\toprule
\rowcolor{Gray}
& & \multicolumn{8}{c}{\textbf{RAG (Prompt \textless{}INS, Q, top-3 Chunks\textgreater{})}} & \multicolumn{8}{c}{\textbf{\fulltext (Prompt \textless{}INS, Q, Full-text\textgreater{})}} \\ 
\rowcolor{Gray}
\textbf{Size} & \textbf{Model} & $\textbf{\texttt{R}}_{\mu}$ & \textbf{\texttt{BL}} & \textbf{\texttt{BS}} & \textbf{\texttt{L70}} & \textbf{\texttt{4o}} & \textbf{\texttt{4oM}} & \textbf{\texttt{ALS}} & \textbf{\texttt{Avg}} & $\textbf{\texttt{R}}_{\mu}$ & \textbf{\texttt{BL}} & \textbf{\texttt{BS}} & \textbf{\texttt{L70}} & \textbf{\texttt{4o}} & \textbf{\texttt{4oM}} & \textbf{\texttt{ALS}} & \textbf{\texttt{Avg}} \\ 
\cmidrule(r){1-2} \cmidrule(lr){3-10} \cmidrule(l){11-18}
\multirow{3}{*}{2-3 B} & Gemma IT \citeyearpar{team2024gemma} & 24.7 & 34.9 & 54.0 & 41.0 & 41.1 & 41.2 & 41.1 & 39.5 & 13.5 & 18.9 & 45.5 & 37.1 & 31.7 & 35.4 & 34.7 & 30.3 \\
& Phi-2 \citeyearpar{2023phi2} & 28.7 & 35.2 & 54.2 & 27.3 & 51.4 & 56.4 & 45.0 & 42.2 & 16.2 & 24.6 & 49.0 & 53.1 & 48.1 & 54.7 & 52.0 & 41.0 \\
& Qwen 2.5 IT \citeyearpar{qwen2.5} & 7.1 & 43.1 & 44.4 & 32.5 & 30.1 & 40.8 & 34.5 & 33.0 & 17.5 & 24.7 & 46.8 & 45.7 & 37.7 & 41.9 & 41.8 & 35.7 \\
\cmidrule(r){1-2} \cmidrule(lr){3-10} \cmidrule(l){11-18}
\multirow{7}{*}{6-7 B} & Falcon IT \citeyearpar{falconmodel} & 23.9 & 37.3 & 53.7 & 45.8 & 49.5 & 55.5 & 50.3 & 44.3 & 18.3 & 28.6 & 51.3 & 49.5 & 50.1 & 55.8 & 51.8 & 42.3 \\
& Galactica \citeyearpar{taylor2022galactica} & 20.7 & 29.3 & 49.8 & 17.3 & 40.6 & 47.9 & 35.3 & 34.3 & 20.5 & 24.9 & 49.8 & 53.1 & 52.3 & 57.8 & 54.4 & 43.1 \\
& Llama 2 Chat \citeyearpar{llama2model} & 24.1 & 35.6 & 53.8 & 56.8 & 53.8 & 57.6 & 56.1 & 47.0 & 15.1 & 30.2 & 52.0 & 59.4 & 56.7 & 62.5 & 59.5 & 46.0 \\
& Llama 3.1 IT \citeyearpar{llama31model} & 5.9 & 51.6 & 42.7 & 40.7 & 39.6 & 51.1 & 43.8 & 38.6 & 16.6 & 30.5 & 51.3 & 57.8 & 57.4 & 60.8 & 58.7 & 45.7 \\
& Longchat 32k \citeyearpar{longchat2023} & 19.8 & 38.1 & 52.6 & 53.1 & 50.6 & 55.7 & 53.1 & 45.0 & 15.8 & 22.9 & 49.5 & 56.4 & 47.8 & 51.1 & 51.8 & 40.6 \\
& Mistral IT \citeyearpar{jiang2023mistral} & 24.4 & 38.4 & 55.2 & 55.9 & 54.1 & 58.0 & 56.0 & 47.7 & 18.6 & 24.4 & 50.4 & 54.3 & 50.5 & 55.6 & 53.5 & 42.3 \\
& Qwen 2.5 IT \citeyearpar{qwen2.5} & 8.9 & 48.4 & 45.7 & 54.3 & 53.5 & 60.0 & 55.9 & 45.1 & 17.0 & 29.3 & 48.0 & 53.8 & 49.5 & 50.9 & 51.4 & 41.4 \\
& Vicuna \citeyearpar{vicuna2023} & 23.8 & 33.3 & 53.2 & 50.2 & 42.7 & 48.9 & 47.3 & 42.0 & 15.8 & 29.0 & 52.3 & 61.2 & 58.1 & 62.3 & 60.5 & 46.5 \\
& Zephyr $\beta$ \citeyearpar{zephyrmodel} & 18.8 & 39.4 & 54.5 & 59.1 & 58.9 & 61.8 & \textbf{59.9} & 48.7 & 16.5 & 24.5 & 50.5 & 56.7 & 49.7 & 54.9 & 53.8 & 42.1 \\
\cmidrule(r){1-2} \cmidrule(lr){3-10} \cmidrule(l){11-18}
\multirow{2}{*}{13 B} & Llama 2 Chat \citeyearpar{llama2model} & 22.0 & 39.0 & 55.3 & 58.5 & 53.1 & 57.3 & 56.3 & 47.5 & 15.4 & 30.1 & 52.0 & 58.2 & 57.4 & 62.2 & 59.3 & 45.9 \\
& Vicuna \citeyearpar{vicuna2023} & 25.6 & 35.4 & 54.6 & 53.7 & 49.5 & 53.6 & 52.3 & 45.4 & 16.0 & 28.2 & 51.9 & 62.1 & 59.5 & 62.9 & 61.5 & 46.8 \\
\cmidrule(r){1-2} \cmidrule(lr){3-10} \cmidrule(l){11-18}
\multirow{2}{*}{70 B} & Llama 2 Chat \citeyearpar{llama2model} & 16.0 & 26.4 & 43.1 & 56.3 & 52.4 & 50.1 & 52.9 & 40.7 & 11.7 & 18.7 & 45.3 & 35.7 & 33.6 & 35.9 & 35.1 & 30.1 \\
& Llama 3.1 IT \citeyearpar{llama31model} & 23.0 & 37.9 & 54.0 & 57.0 & 55.5 & 58.2 & 56.9 & 47.6 & 18.2 & 28.9 & 52.7 & 61.4 & 61.6 & 63.9 & 62.3 & 47.8 \\
\cmidrule(r){1-2} \cmidrule(lr){3-10} \cmidrule(l){11-18}
\multirow{4}{*}{UNK} & Gemini Pro \citeyearpar{team2023gemini} & 24.2 & 30.2 & 49.9 & 42.0 & 41.5 & 40.4 & 41.3 & 38.0 & 6.9 & 27.5 & 42.9 & 51.3 & 48.5 & 48.4 & 49.4 & 37.6 \\
& GPT-4o & 28.5 & 36.5 & 55.6 & 51.9 & 53.4 & 53.9 & 53.1 & 46.6 & 26.2 & 40.3 & 57.4 & 66.2 & 66.6 & 67.5 & 66.8 & \textbf{54.0} \\
& GPT-4o-mini-2 & 25.4 & 38.0 & 56.0 & 57.2 & 57.9 & 58.9 & 58.0 & \textbf{48.9} & 22.4 & 40.8 & 56.9 & 68.0 & 67.2 & 68.8 & 68.0 & \textbf{54.0} \\ 
& GPT-4 & - & - & - & - & - & - & - & - & 10.0 & 33.1 & 48.4 & 68.2 & 67.6 & 68.5 & \textbf{68.1} & 49.3 \\ \bottomrule
\end{tabular}
}
\end{center}
\caption{RAG setup prompts the LLM with top-3 chunks extracted from the paper. \fulltext evaluation - LLMs are provided with the paper's full-text. If the full-text exceeds LLM's context length, the base-LLM reasons over paper chunks and generates answer candidates, followed by Llama 3.1 70B for answer selection.}
\label{tab:ragcft_results}
\end{table*}

\paragraph{Overall high score may not correlate with reasoning from the context.}
GPT-4o and GPT-4o-mini models perform the best among all evaluated models, and achieve the highest average score in the \fulltext setting. However, the GPT models also perform best in both the \closedbook settings, which indicates that the model is also able to reason about the questions without providing the context. Priming with the paper title and abstract (\titleabs) leads to 2-3 points improvement over the full \closedbook setting for both models. In comparison to \titleabs, the average score improves by four points in \fulltext setting. This might indicate the model is able to retrieve relevant knowledge to the question from its parameters without explicitly being provided with the paper, in which case the model had been trained on the source papers.\footnote{This also might suggest potential for contamination affecting the results. Although our question and answers are revised and rewritten, there's a chance that training on raw open-review data might help the models in this task.}

Among the open-source LLMs, the best scores are achieved by Llama 3.1 70B Instruct model, however, its average score for \fulltext and RAG setup is within one point difference of other open-domain question answering setups, which means the model is not using the provided context from the paper for reasoning about the questions. Other models with performances similar to Llama 3.1 70B, and also showcasing significant improvements over the corresponding \closedbook settings are Vicuna 7B, 13B and Llama 2 13B Chat. 

Instruction-tuned models perform better than their counterparts generally. The instruction-tuned counterparts of Gemma, Falcon, Llama 2, and Mistral perform better at retrieving the answers.

\paragraph{Overall, GPT-4o and GPT-4o-mini perform best among all evaluated models in all task settings. }
For \closedbook and \titleabs task settings, we report the scores with surface-level metrics, LLM judge scores and overall average scores in~\Cref{tab:odqa_results}. GPT-4o and GPT-4o-mini perform best among evaluated models in both settings. Among open-source models, Llama 3.1 70B Instruct and Qwen2.5 7B Instruct models perform the best in both \closedbook and \titleabs settings. In RAG setting, Zephyr $\beta$ 7B and GPT-4o-mini perform the best, followed by similar performances from GPT-4o, Llama 3.1 8B Instruct, Llama 2 Chat 13B, Llama 2 Chat 7B, Mistral 7B Instruct, and Qwen2.5 7B Instruct. However, the best overall score in RAG setting (\Cref{tab:ragcft_results}) is similar to \closedbook and \titleabs settings, indicating that providing the context chunks does not lead to significantly higher scores. 

\paragraph{Best performance is observed in \fulltext setting, with significant differences among scores of proprietary and open models.}
For the full-text setting (\Cref{tab:ragcft_results}), a significant score difference is observed among proprietary models (GPT-4o, GPT-4o-mini) and best-performing open-source models (Llama 3.1 70B Instruct and Vicuna 13B).

\paragraph{Human Performance Estimation:} 
Evaluating human performance on the SciDQA dataset is challenging due to the complex and domain-specific nature of its questions. To assess human proficiency, the authors compared human-written responses from 28 annotated QA pairs against those generated by GPT-4.\footnote{GPT-4 shows simlilar performance to GPT-4o in our LLM judge metrics, and this experiment was done earlier than GPT-4o's release.} An author performs the task by writing the answers to the given questions, by reading and examining the paper. The annotator found this task to be relatively challenging, particularly for papers outside their expertise. During evaluation, each instance included a question, a ground truth answer, an author-written answer after reading the paper, and a GPT-4 generated answer; with evaluations focusing on comprehensiveness, factuality, and clarity. Results showed that 32\% of comparisons ended in a tie, indicating GPT-4’s adequacy for simpler questions. Humans were preferred in 29\% of cases, mainly due to factual inaccuracies in GPT-4 responses. GPT-4 outperformed humans in 21\% of instances; these cases were mostly related to papers whose topics were outside the authors' expertise. However, 18\% of both answers were rejected as unsatisfactory, particularly for complex questions. Detailed performance metrics are available in the Appendix~\Cref{tab:scores_human_gpt4}. 

\section{Related Work}
\paragraph{Manually Curated Scientific QA Datasets:} The QASPER dataset~\citep{dasigi2021dataset} involves NLP practitioners creating questions from paper titles/abstracts, with answers derived from full-texts by separate annotators. The QASA dataset~\citep{lee2023qasa} is generated by AI/ML practitioners and paper authors who formulate surface, testing, and deep questions. In contrast, the COVID-QA dataset~\citep{moller-etal-2020-covid} is crafted by 15 biomedical experts, who develop questions and annotate corresponding text as answers, focusing on COVID-19 research. QASPER has 40\% questions answered in less than five words, while 30\% of QASA QA pairs are sourced from only the introductions and abstracts, with 52\% of answers showing high unigram overlap with the text, indicating easier retrieval. 
The ExpertQA dataset~\citep{malaviya2024expertqa} features 2,177 questions across 32 fields, created by 524 experts to simulate complex, web-based information-seeking scenarios. BioASQ-QA dataset~\citep{krithara2023bioasq,tsatsaronis2015overview} involves expert-curated questions ranging from yes/no, factoid, list, and summary types, growing from 310 to 4,721 instances over ten years. Since 2016, BioASQ-QA has focused solely on titles and abstracts, reflecting the high effort in manual curation. 

\paragraph{Synthetically Generated Scientific QA Datasets:} 
BioRead~\citep{pappas-etal-2018-bioread} and BioMRC~\citep{pappas-etal-2020-biomrc} are cloze-style biomedical MRC datasets that utilize text entities as answers, masking these entities in texts (passages in BioRead, abstracts in BioMRC) and forming questions from the last passage line or title. ScholarlyRead~\citep{saikh-etal-2020-scholarlyread} generates QA pairs by extracting noun phrases from abstracts and using a question-generation model.  As shown in~\Cref{tab:data_stats}, these synthetically generated QA datasets generally feature shorter answers than ours. PubMedQA~\citep{jin-etal-2019-pubmedqa} starts with a labeled dataset where the title is a question and the last abstract line is the answer, creating 1000 instances with short answers (yes/no/maybe) annotated based on the abstract. Its synthetic counterpart uses syntax heuristics and modification rules to craft similar QA pairs.

\paragraph{Other datasets:} 
The ARIES dataset~\citep{d2023aries} compiles review comments and associated paper edits. Its synthetic subset uses a method similar to ours to extract comment-edit pairs based on textual overlap. Our dataset diverges by extracting questions from review comments using LLMs, not just from quoted responses but also from author rewrites. We employ human-expert annotation to refine questions and answers, avoiding reliance solely on textual overlap. This allows us to include high-quality queries involving tables, equations, and multi-paragraph reasoning. In a parallel direction,~\citet{kang-etal-2018-dataset} collect peer review datasets for paper acceptance prediction and score prediction for review aspects tasks.

\Dname stands out among QA datasets as its questions are sourced directly from the peer review process, ensuring they are natural, evaluative, and of high quality due to the scientific discourse among domain experts. This sourcing ensures that the questions require a deep understanding of the content, emphasizing depth as well as quality. 

\section{Conclusion and Future Work}
We curate \Dname, dataset designed to challenge language models on the QA task aiming to evaluate their understanding of scientific papers. The dataset consists of 2937 QA pairs, and extracts QA asked by reviewers and answered by paper authors during reviewer-author discussion during manuscript review on OpenReview. Our multi-stage refinement pipeline annotates for quality, decontextualizes the QA pairs, edits references, and establishes the source document from different manuscript versions. Our dataset features questions necessitating reasoning across multiple modalities beyond mere text, including figures, tables, equations, appendix and supplementary materials. \Dname also provides a testbed for evaluation of multi-document comprehension properties of LLMs. We evaluate the performance of several open-source and proprietary LLMs in generating the answer to questions after comprehending the research paper. We posit that \Dname will serve as a useful resource to benchmark the performance of LLMs in scientific text comprehension.

\section{Limitations}
Multiple questions in our dataset necessitate comprehension and reasoning over multiple documents. The questions in the dataset often mention the reference text for previous works that need to be referred to for answering the question. However, in our experiments we do not search and include those documents for answer generation. Additionally, 7\% questions are dependent on figures, but the Nougat parser does not extract images and only extracts the figure captions. We do not evaluate any visual or multimodal LLM. We extract figures for the specific figure-related questions using PDFFigures~\citep{pdffigures}, summarize it using Llama 3.2 and make that available. Large-scale evaluation of free-form generation is still challenging. We provided both surface-level and LLM-as-a-judge metrics to show the full picture of the performance, however, extensive meta-evaluation studies might be needed to carefully understand the limitations of such metrics in our setting. Additionally, the dataset could be used to generate difficult questions from a manuscript. Our dataset does not have any judgment statements about paper acceptance/rejection. However, the questions dataset could still be used for training a question generator, and complex questions could be misused by reviewers as grounds for rejection. Another limitation is disentangling the effect of potential contamination from performance of various evaluated models, which is difficult to do for models that don't discuss their training data (which includes majority of both closed and open weight models). Finally, similar to other existing datasets, our dataset focuses on curating QA pairs from a specific domain (machine learning), rather than all scientific fields of study.

\section*{Acknowledgments}
We extend our gratitude to Mike D'Arcy for providing feedback and valuable discussion on the paper. We also thank the anonymous reviewers and area chairs for their feedback. We are grateful for the compute support provided by the Microsoft Research’s Accelerate Foundation Models Research (AFMR) program and Google's TRC program. Shruti is grateful for the support received through the Fulbright-Nehru Doctoral Research Fellowship to visit Yale University. Shruti is also supported by the Prime Minister’s Research Fellowship (PMRF-1701251) awarded by the Government of India.

\bibliography{anthology,custom}

\appendix

\section{Curation of \Dname - Data Pre-processing, Annotation and Editing}
\label{app:preproc}
\paragraph{Curation from OpenReview}
We selected top-tier machine learning and deep learning venues, designated as A* rankings by ICORE Portal \citep{icoreranking}, with publicly accessible reviewer-author discussions on OpenReview. 
During the dataset compilation phase, NeurIPS, ICLR, ICML, and TMLR were the A* venues with available discussions. However, only discussions from ICML workshop papers and accepted papers from TMLR were accessible, with rejected papers from TMLR not being included. To ensure high quality, we excluded ICML workshop papers. Further, TMLR was also excluded to maintain diversity and avoid a narrow focus on only accepted papers. We curate 11400 papers from ICLR (2018-2022) and NeurIPS (2021-2022), with major focus to include newer papers to decrease the risk of contamination with LLM pretraining datasets.

\paragraph{PDF to Text Conversion}
OpenReview portal hosts the multiple versions of PDF files for papers submitted to ICLR and NeurIPS, which also includes the versions uploaded during the discussion phase. We downloaded the last manuscript submitted prior to the conference deadline, and refer to it as the initial version, as well as the final manuscript, known as the camera-ready version. In case of rejected manuscripts, the camera-ready version is not uploaded, and hence, we either take the latest version submitted during discussion with reviewers, or take the initially submitted manuscript. For converting PDFs to text, we employed Nougat~\citep{blecher2023nougat}, a visual transformer model designed for the optical character recognition (OCR) task. Nougat parses research paper PDFs into markdown format and has been trained on a dataset comprising papers from arXiv, Pubmed Central, and the Industry Document Library. We opted for Nougat as it is the current state-of-the-art, showcasing superior performance in extracting tables, mathematical text (equations), and general text compared to GROBID~\citep{GROBID}, another widely used OCR tool.

\paragraph{Regex Filtering}
OpenReview has nested discussions, i.e. authors and reviewers reply to corresponding messages, creating a time-stamp chain of discussion. Reviewers post the initial review message, generally consisting of paper summary, strengths and weakness, questions to authors, and a recommendation score. Segments of reviewer messages may be quoted in markdown or paraphrased by the authors in their replies to address specific content. Subsequently, reviewers may ask additional clarifying questions based on the authors' responses, or express satisfaction or dissatisfaction. There are instances where, despite the reviewers' questions, the authors do not provide responses. To extract nested discussions containing at least one question and answer, we employed regex pattern matching, searching for cues such as `Question:', `Q', etc. Using this method, we extracted 18,658 reviewer-author discussions for 11,400 papers that contained questions and answers.
We use the following regex pattern to identify discussions that contain some questions:
\begin{tcolorbox}[enhanced,
  colbacktitle=Gray, title=Regex for Extraction,fonttitle=\bfseries,coltitle=black,
  boxed title style={size=small,colframe=black,colframe=black} ]
\begin{verbatim}
"que[ 0-9]*?[:-] .*[^\n]"
"Q[ 0-9]*?[:-] .*[^\n]"
"question[ 0-9]*?[:-] .*[^\n]"
"^> .*[^\n]"
\end{verbatim}
\end{tcolorbox}

\paragraph{LLM-based QA Extraction}
The prompt provided to PaLM text-bison-001 model to extract QA pairs is presented in~\Cref{fig:PALMExtractionPrompt}.

\begin{figure}
    \begin{tcolorbox}[enhanced,attach boxed title to top center={yshift=-3mm,yshifttext=-1mm},
      colbacktitle=Gray, title=Prompt for QA Extraction using PaLM,fonttitle=\bfseries,coltitle=black,
      boxed title style={size=small,colframe=black,colframe=black} ]
    
    You are a helpful assistant. Read the following paragraph and find all question-answer pairs in it. \\
    \\
    \textit{Author Response to Reviewer}\\
    \\
    Add `Q:' before each question and `A:' before answers. The question-answer pairs are: 
    \end{tcolorbox}
    \caption{Prompt for PaLM model to extract question-answer pairs from Reviewer-Author discussions.}
    \label{fig:PALMExtractionPrompt}
\end{figure}

\begin{figure*}[!tbp]
    \centering
    \includegraphics[width=0.9\linewidth]{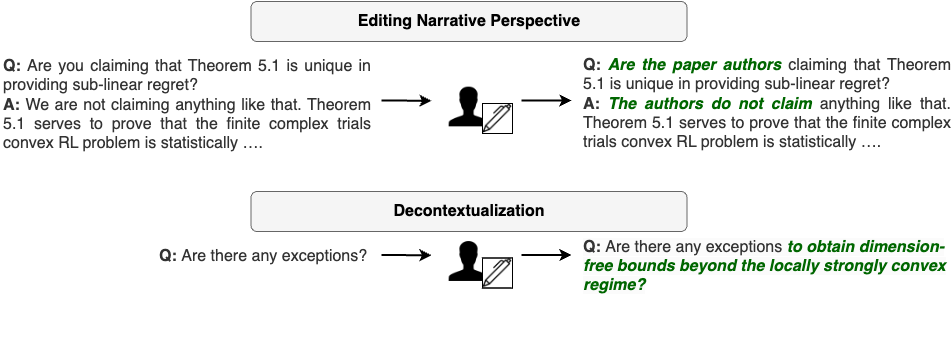}
    \caption{Rewriting QA pairs in a third-person narrative is crucial for models to recognize that questions seek factual answers based on the author's reasoning in the paper, rather than personal opinions. Furthermore, incorporating contextual information enhances the comprehension of questions that necessitate prior contextual knowledge for accurate interpretation.}
    \label{fig:editing_qa}
\end{figure*}
\begin{figure*}[!h]
    \centering
    \includegraphics[width=0.9\linewidth]{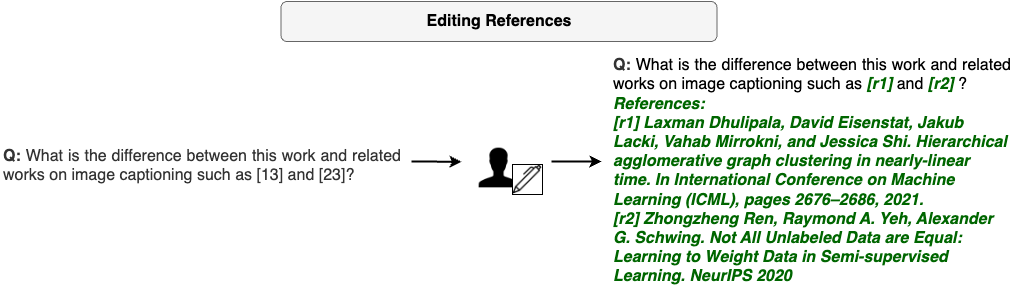}
    \caption{References in question and answer texts are uniformly renumbered (e.g., r1, r2, or 1, 2, or A, B) to preclude the LM from leveraging specific reference markers as shortcuts for answer retrieval. To facilitate accurate answer formulation by the LM, textual information pertaining to paper references is incorporated into questions, deterring reliance on mere reference numbers. Similarly, references in answers are renumbered and supplemented with the relevant reference text as necessary.}
    \label{fig:ref_editing}
\end{figure*}

\subsection{Annotation details}
\label{apx:annotation}
The annotators achieved an 85\% agreement rate in filtering the type of questions as relevant, irrelevant or ambiguous. Half of the disagreements pertained to the ambiguous category, with discrepancies arising from one annotator marking instances as ‘ambiguous’ to speed up annotation versus another favoring detailed assessment. In such cases, the annotation disagreement does not imply disagreement regarding the inclusion of the instance in the dataset. The annotators resolved the remaining disagreements through discussion and refined the annotation guidelines to eliminate ambiguities before proceeding with the rest of the dataset. 

The annotation process was facilitated by providing details such as the paper title, submission venue, area chair recommendations, and the extracted questions with their corresponding answers. To minimize the workload, questions from the same paper but different reviewers were assigned to the same annotator. Annotators were encouraged to consult the original review texts for additional context, enhancing the accuracy of their annotations. Some instances of QA pairs that are marked as relevant, irrelevant, or ambiguous are presented in~\Cref{tab:cat_instances}.

We present scenarios depicting the requirement of editing QA pairs, and the references text to improve dataset quality in~\Cref{fig:editing_qa} and~\Cref{fig:ref_editing}.

\begin{figure}[!hp]
\centering
\begin{subfigure}[b]{.46\textwidth}
   \includegraphics[width=0.9\linewidth]{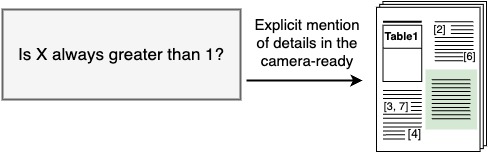}
   \caption{Initial version is preferred as the revised copy explicitly mentions the answer.}
\end{subfigure} \\
\begin{subfigure}[b]{.46\textwidth}
   \includegraphics[width=0.9\linewidth]{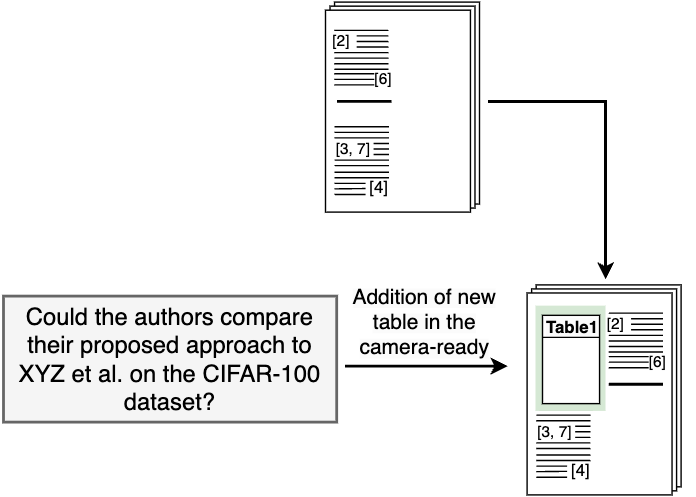}
   \caption{Revised version is preferred as the answer is originally absent.}
\end{subfigure}
\caption{We present scenarios where the initial and the revised manuscript versions are most appropriate for answering the reviewer's question. For each question in the dataset, we annotate the preferred manuscript version.}
\label{fig:initial_vs_camready}
\end{figure}

\paragraph{Source Document Annotation}
Scenarios depicting cases where initial vs revised manuscripts are appropriate for answering the reviewer questions are presented in~\Cref{fig:initial_vs_camready}.

\paragraph{Evidence Extraction}
We extract evidential paragraphs, figures, tables, and lines in paper text from the author responses. We also extract evidences for a smaller subset of the dataset automatically where there is a high overlap between a paper section and the answer.

\newcolumntype{A}{>{\centering\arraybackslash} p{0.95\linewidth}}
\newcolumntype{B}{p{0.95\linewidth}}
\begin{table*}[!hbp]
\centering
\small{
\begin{tabular}{A}
\hline
\rowcolor{Gray}
\multicolumn{1}{c}{Relevant for \Dname} \\ \hline
\begin{tabular}[c]{@{}B@{}}Q: How is the expectation of TCE algorithm computed in Equation 18?\\ A: The expectation is calculated with respect to the ...\end{tabular}   \\
\begin{tabular}[c]{@{}B@{}}Q:  In section 3.4.1, is it possible to apply ReMERT to non-episodic or continuing task?\\ A: ReMERT might not provide proper weights to ….\end{tabular} \\
\hline
\rowcolor{Gray}
\multicolumn{1}{c}{Irrelevant for \Dname} \\ \hline
\begin{tabular}[c]{@{}B@{}}Q: Can the inversion method by Chen et al. 2022 be used to improve the latency?\\ A: We believe that this may be possible, however it will require further analysis.\end{tabular} \\
\begin{tabular}[c]{@{}B@{}}Q: Can you correct the typos in Section 3.4?\\ A: Yes, we will correct them in the revised version.\end{tabular}  \\
\hline
\rowcolor{Gray}
\multicolumn{1}{c}{Ambiguous}  \\ \hline
\begin{tabular}[c]{@{}B@{}}Q: Can this inversion method be used in tandem with online filtering/smoothing (e.g. 4DVar, EnKF)?\\ A: We believe that this may be possible, potentially leveraging ideas from Chen et al. {[}2021{]}.\end{tabular}            \\ \hdashline
\begin{tabular}[l]{@{}B@{}}Q: Why don't the authors compare to PINNs?\\ A: PINNs are typically employed to retrieve individual solutions, not learn distributions over data sets. When using them to solve the individual problems, inference is much slower since the network needs to be trained for each inferred solution. Iterative solvers seem like a better alternative in our setting.\end{tabular} \\ \hline
\end{tabular}
}
\caption{Categorization of questions for inclusion in the \Dname~dataset. Information-seeking questions, whose answers are ascertainable within the research paper text, from a collection of synthetically extracted question-answer pairs using PaLM text-bison-001 model are categorized as relevant, and added to the \Dname~dataset.}
\label{tab:cat_instances}
\end{table*}

\begin{figure}[!htbp]
    \centering
    \includegraphics[width=0.9\linewidth]{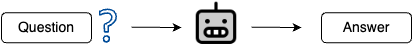}
    \caption{Priming LLMs with Questions (\closedbook). This task evaluates the ability of LLM to recall the answer without any relevant context.}
    \label{fig:mem}
\end{figure}

\begin{figure}[!htbp]
    \centering
    \includegraphics[width=0.9\linewidth]{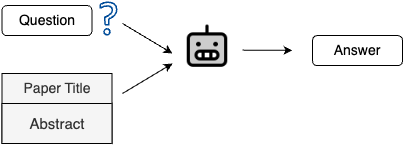}
    \caption{Open-Domain Question Answering - Priming with Question and Title/Abstract (\titleabs). This task evaluates the impact of additional context on LLM ability to recall the answer without reasoning about the question.}
    \label{fig:tabs}
\end{figure}
\begin{figure*}[!t]
    \begin{minipage}{0.98\textwidth}
        \centering
        \includegraphics[width=0.8\linewidth]{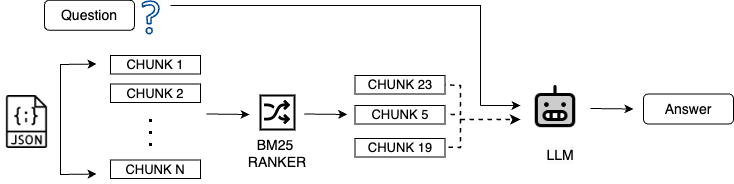}
        \caption{RAG setup ranks paper subsections based on their relevance to the question, and top-3 subsections are provided to the base-LLM, which generates the answer.}
        \label{fig:rag}
    \end{minipage}
    \hfill
    \begin{minipage}{0.98\textwidth}
        \centering
        \includegraphics[width=0.9\linewidth]{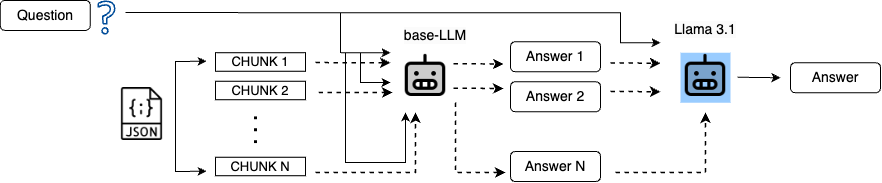}
        \caption{Comprehending the full-text (\fulltext) of the research paper by passing model context-length segments to the base-LLM and generating answers from each segment. Llama 3.1 70B selects the best answer among generated candidate answers. For models with 128k context length, such as Gemini, GPT-4o, and GPT-4o-mini, the entire text is provided to the base-LLM in a single chunk and answer selection phase is not required.}
        \label{fig:ft}
    \end{minipage}
\end{figure*}

\section{Experiments}
\label{app:exp_details}
\subsection{Experimental Setup}
We present figures for the experimental setup of the following:
\begin{enumerate}[noitemsep,nolistsep]
    \item Open-Domain Question Answering - Priming with Questions (\closedbook) in~\Cref{fig:mem}.
    \item Open-Domain Question Answering - Priming with Question and Title/Abstract (\titleabs) in~\Cref{fig:tabs}.
    \item Retrieval Augmented Generation (RAG) in~\Cref{fig:rag}.
    \item Comprehending the full-text (\fulltext) in~\Cref{fig:ft}. The figure presents the scenario where the full-text cannot fit into the models' context length. 
\end{enumerate}

We experimented with the parameters (temperature=0.1, 0.9, top\_p=0.1, 0.5, 0.9) on a smaller subset of 20 QA pairs, and selected temperature=0.1 and top\_p=0.9 after manually inspecting LLM answers. We carried out three runs initially, but upon observing no significant difference in performance, we reported the final numbers in the paper using a single run.

\subsubsection{Chunk Creation Algorithm}
\paragraph{Chunking for RAG Setup}
RAG setup ranks top-k chunks from the full-text which are then provided to the LLM to generate the answer. The chunking strategy is presented in~\Cref{alg:chunking_rag}, and ensures that the individual chunks fit into the model context length. It also ensures that the collective top-k chunk lengths also fit the model context length, and sentences from different paper sections or paragraphs are not collated together in a single chunk. We found this setting to work better than naive chunking and truncating by paragraphs or sections.

\paragraph{Chunking for FT Setup}
In the \fulltext setting, the chunk length is determined by the LLM’s context length. If the model context length is N, we reserved 500 tokens for the instruction and the question, and utilized the rest N - 500 tokens for context. The chunking strategy is presented in~\Cref{alg:chunking_ft}.

For Gemini, GPT-4o, GPT-4o-mini, and GPT-4; chunking is not required and the answer selection phase is not included in final answer generation. For models with 128k context-length, Qwen v2.5 models (1.5B and 7B) and Llama 3.1 70B models (8B and 70B), the prompt with entire full-text does not fit on our GPUs, so we create chunks as it is done with other smaller context-length LLMs. However, for \fulltext generations with Qwen v2.5 and Llama 3.1 models, the base-LLM is used for final answer selection. With other base-LLMs, all generated candidate answers concatenated together exceed the context-length of base-LLM so Llama 3.1 70B is used for final answer selection.

\begin{algorithm}
\caption{Chunk Creation Algorithm for RAG}
\label{alg:chunking_rag}
\begin{algorithmic}[1]
\State \textbf{Input:} Full-text document
\State \textbf{Output:} List of chunks
\State Split the full-text into paragraphs (demarcated by \textbackslash n).
\For{each paragraph $P$}
    \State Split $P$ into individual sentences (using the NLTK library).
    \State Initialize an empty list $chunks$
    \For{every 10 consecutive sentences in $P$}
        \State Join the sentences to build a chunk.
        \State Add the chunk to $chunks$
        \State Slide the window by nine sentences (i.e., keep a single overlapping sentence between consecutive chunks).
    \EndFor
\EndFor
\end{algorithmic}
\end{algorithm}

\begin{algorithm}
\caption{Chunk Creation Algorithm for \fulltext}
\label{alg:chunking_ft}
\begin{algorithmic}[1]
\State Split the full-text into paragraphs (demarcated by \textbackslash n).
\For{each paragraph}
    \If{the length of paragraph is less than $N - 500$}
        \State The entire paragraph is treated as a chunk
    \Else
        \State Split the paragraph into a list of sentences, say $S = [s_1, s_2, \dots, s_n]$
        \State Initialize an empty $chunks\_list = [\ ]$
        \State Initialize an empty string chunk $c = ``"$
        \For{sentence $s$ in $S$}
            \If{$token\_count(c) + token\_count(s) < N - 500$}
                \State Add sentence $s$ to the chunk $c$
                \State \textbf{continue}
            \Else
                \State Add chunk $c$ to the $chunks\_list$
                \State Reinitialize the empty chunk $c$
                \State Add sentence $s$ to the chunk $c$
                \State \textbf{continue}
            \EndIf
        \EndFor
    \EndIf
\EndFor
\end{algorithmic}
\end{algorithm}

\subsection{Answer Selection Prompt for Llama 3.1 70B}
The prompt provided to Llama 3.1 70B model to generate a single answer during the answer selection phase in \fulltext setup is presented in~\Cref{fig:answer_selection_prompt}.
\begin{figure}
    \begin{tcolorbox}[enhanced,attach boxed title to top center={yshift=-3mm,yshifttext=-1mm},
      colbacktitle=Gray, title= Answer Selection Prompt - Llama 3.1,fonttitle=\bfseries,coltitle=black,
      boxed title style={size=small,colframe=black,colframe=black} ]
      \label{box:promptgemini}
      You are provided with a question and some potential answers about a research paper submitted to a top-tier computer science conference in the domain of ML and DL. Your task is to select the best answer from the provided answer options, which comprehensively answers the question. Do not include any additional text other than the answer and select only one answer from the provided options.
    \end{tcolorbox}
    \caption{Prompt provided to Llama 3.1 70B to select one candidate answer among multiple candidates generated from multiple chunks.}
    \label{fig:answer_selection_prompt}
\end{figure}

\subsection{LLM Judge Prompts}
\label{appx:llm-judge-prompts}
The prompt provided to models to generate an evaluation of relevance, accuracy, completeness, and conciseness aspects is presented in~\Cref{fig:promptLLMJudge}. The prompt provided to the Llama 3.1 8B model to extract the overall score from the evaluation statement is presented in~\Cref{fig:score_extraction}.

\begin{table}[htbp]
\centering
\newcolumntype{C}{p{0.2\linewidth}}
\small{
\begin{tabular}{Ccccc}
\toprule
\rowcolor{Gray}
Preferred Answer $\rightarrow$ / Score $\downarrow$ & Tie  & Human & GPT-4 & None \\ 
\midrule
GPT-4 & 32.5 & 30.4  & 37.0  & 34.6 \\
Human & 34.8 & 34.4  & 38.5  & 34.0 \\ \bottomrule
\end{tabular}}
\caption{Average scores of Human and GPT-4 generated answers on a subset of SciDQA dataset across instance categories. The average score (R-1, R-2, R-L, \texttt{BL}, \texttt{BS}) of human and GPT-4 generated answers are grouped by preference.}
\label{tab:scores_human_gpt4}
\end{table}

\begin{figure*}
\begin{tcolorbox}[enhanced,attach boxed title to top center={yshift=-3mm,yshifttext=-1mm},
  colbacktitle=Gray, title=Prompt for LLM Judge,fonttitle=\bfseries,coltitle=black,
  boxed title style={size=small,colframe=black,colframe=black},width=\textwidth ]
You are an expert evaluator tasked with assessing the quality of a model-generated answer compared to a gold standard correct answer in a long-form question-answering context. Your goal is to provide a quantified evaluation across multiple dimensions. Please follow these steps:
\\
Carefully read the original question, the model-generated answer, and the gold correct answer. Evaluate the model-generated answer on the following dimensions, providing a score from 1-10 for each (where 1 is poor and 10 is excellent): a) Relevance (1-10): How well does the answer address the specific question asked? b) Accuracy (1-10): To what extent is the information provided correct and aligned with the gold answer? c) Completeness (1-10): How thoroughly does the answer cover all aspects of the question compared to the gold answer? d) Conciseness (1-10): Does the answer provide information efficiently without unnecessary details?
\\
Calculate an overall quality score by taking the average of the five dimension scores. In your answer for each dimension, provide a justification why not a higher score and why not a lower score.
\\
Structure your response as follows:
\\
Evaluation:\\
1. Relevance: [Score] - [Explanation]\\
2. Accuracy: [Score] - [Explanation]\\
3. Completeness: [Score] - [Explanation]\\
4. Conciseness: [Score] - [Explanation]\\
\\
Overall Quality Score: [Average of the four above scores]
\end{tcolorbox}
\caption{Prompt provided to Llama 3.1 70B model during answer selection phase in \fulltext.}
\label{fig:promptLLMJudge}
\end{figure*}

\begin{figure*}
\begin{tcolorbox}[enhanced,attach boxed title to top center={yshift=-3mm,yshifttext=-1mm},
  colbacktitle=Gray, title=Prompt for Extraction of Scores from LLM Evaluation Statements,fonttitle=\bfseries,coltitle=black,
  boxed title style={size=small,colframe=black,colframe=black},width=\textwidth ]
You are provided with an evaluation of an answer in the following format: \\
\\
Evaluation:\\
1. Relevance: [Score] - [Explanation]\\
2. Accuracy: [Score] - [Explanation]\\
3. Completeness: [Score] - [Explanation]\\
4. Conciseness: [Score] - [Explanation]\\
Overall Quality Score: [Average of the four above scores].\\
\\
Carefully read the evaluation provided next, and extract the final overall quality score from the discussion. Do not include any explanation, you should only provide the final numeric score for overall quality from the evaluation statement.
\end{tcolorbox}
\caption{Prompt to extract the final overall quality score from the evaluation statements generated by LLM Judges.}
\label{fig:score_extraction}
\end{figure*}

\subsection{Comparison with human-written answers}
\label{appx:human-eval}
Table \ref{tab:scores_human_gpt4} demonstrates the comparison of human-written answers with GPT-4. We present the average scores of surface-level metrics for GPT-4 answers and human-written answers, which are further grouped by categories that indicate which answer was preferred. 

\end{document}